# URBAN-i: From urban scenes to mapping slums, transport modes, and pedestrians in cities using deep learning and computer vision

Mohamed R. Ibrahim[1], James Haworth[2] and Tao Cheng[3]
Department of Civil, Environmental and Geomatic Engineering, University College London (UCL)
[1]mohamed.ibrahim.17@ucl.ac.uk, [2]j.haworth@ucl.ac.uk, [3]tao.cheng@ucl.ac.uk

*Abstract*—Within the burgeoning expansion of deep learning and computer vision across the different fields of science, when it comes to urban development, deep learning and computer vision applications are still limited towards the notions of smart cities and autonomous vehicles. Indeed, a wide gap of knowledge appears when it comes to cities and urban regions in less developed countries where the chaos of informality is the dominant scheme. How can deep learning and Artificial Intelligence (AI) untangle the complexities of informality to advance urban modelling and our understanding of cities? Various questions and debates can be raised concerning the future of cities of the North and the South in the paradigm of AI and computer vision. In this paper, we introduce a new method for multipurpose realistic-dynamic urban modelling relying on deep learning and computer vision, using deep Convolutional Neural Networks (CNN), to sense and detect informality and slums in urban scenes from aerial and street view images in addition to detection of pedestrian and transport modes. The model has been trained on images of urban scenes in cities across the globe. The model shows a good validation of understanding a wide spectrum of nuances among the planned and the unplanned regions, including informal and slum areas. We attempt to advance urban modelling for better understanding the dynamics of city developments. We also aim to exemplify the significant impacts of AI in cities beyond how smart cities are discussed and perceived in the mainstream. The algorithms of the URBAN-i model are fully-coded in Python programming with the pre-trained deep learning models to be used as a tool for mapping and city modelling in the various corner of the globe, including informal settlements and slum regions.

1. INTRODUCTION

Understanding the dynamics of cities within the current global urban challenges- rapid urbanizations, the changeable size of cities, and the high degree of informality and uncertainty- remains a complex process (Batty, 2008; Luís Bettencourt, 2013; Bettencourt & West, 2010), which limits the ability of diagnosing and transferring knowledge from one city to another. Accordingly, finding a reliable method with a unified input data that can enrich the top-down urban strategies in analysing and automating decisions in a wide spectrum of cities across the globe is indeed in high demand; for urban scholars, planners, and policy-makers.

Building our knowledge about cities through images is not a new concept. In fact, Lynch (1960) introduced how to perceive and understand cities from features– landmarks, focal points, skyline, pedestrian flow etc.– that provides a very effective approach to perceiving the nuances of the urban world. It is only more recently that urban scholars have turned to pure mathematical models and theoretical physics related theories to understand the complexity of cities (Batty, 2008; Bettencourt, 2013; Isalgue, Coch, & Serra, 2007).

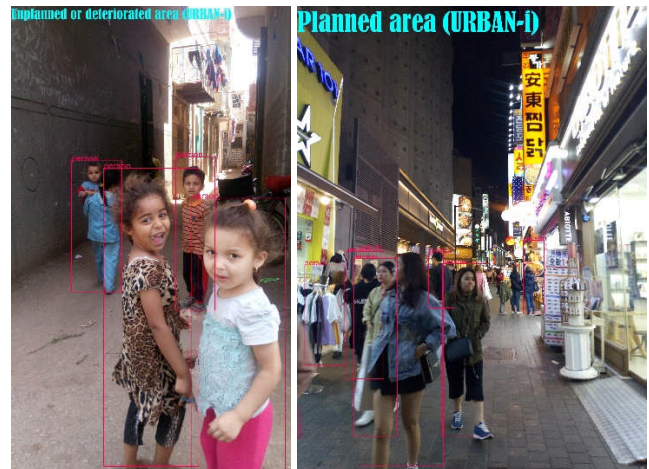

FIG. 1
THE MODEL OUTPUT

While our knowledge of the dynamics of cities is still limited, substantial urban models have been achieved by perceiving cities as complex systems. Various studies have described how to understand cities through cellular automata (Batty, 1997; Batty, Couclelis, & Eichen, 1997; de Almeida et al., 2003), fractals (Batty & Longley, 1994; Batty & Xie, 1996; Murcio, Masucci, Arcaute, & Batty, 2015), or multi-agent modelling (Batty, 2005; Heppenstall, Crooks, See, & Batty, 2012), relying on complexity and network theories. However, due to the complexity of urban reality, the produced models, in many cases, either tend to over-simplify the initial settings of urban systems or rather explore cities in a mono-dimensional perspective (Batty & Torrens, 2001).

On the other hand, geo-tagging and analysing labelled images from social media or any big data providers in cities is another key approach for urban scholars to understand cities (Crandall, Backstrom, Huttenlocher, & Kleinberg, 2009; Gallagher, Joshi, Yu, & Luo, 2009; Hays & Efros, 2015; Quack, Leibe, & Van Gool, 2008; B. Zhou, Liu, Oliva, & Torralba, 2014). Most recently, different attempts have been taken to better understand the geolocations and the spatial structure from urban scenes and image query. While this emerging approach of dealing with urban big data seems promising, the current outputs of such methods in analysing cities, apart from the limited access of the data, is rather used as a data-driven approach that may or may not provide new evidence of the urban structure (Zhou et al., 2014), instead of focusing on understanding the dynamics of the urban systems.

Yet, Lynch's ideas concerning a city's features, in their original forms, may not cope with today's rapid urban challenges. With the growth of the field of deep learning and computer vision, understanding cities through the eyes of a computer opens the door for analysing various missing attributes of city dynamics. Large-scale analysis of digital images and patterns of captured features that may not necessarily be recognized of significance by human eyes can potentially enable various urban issues to be tackled. The accuracy of such models may need further development at an initial stage. However, their advancements appear in keeping track of information and extracting elements from images similar to how urban scholars used to perceive cities. Yet, when it comes to urban development due to the complexity and emerging nature of these models, the produced algorithms are often limited towards the notion of smart cities or autonomous vehicles where the current issues of 'traditional cities', where people actually live, are often undermined. For instance, computer vision was seminal for security-related issues in smart cities (García, Meana-Llorián, G-Bustelo, Lovelle, & Garcia-Fernandez, 2017), and vehicle plate recognition in urban scenes (Tarigan, Nadia, Diedan, & Suryana, 2017).

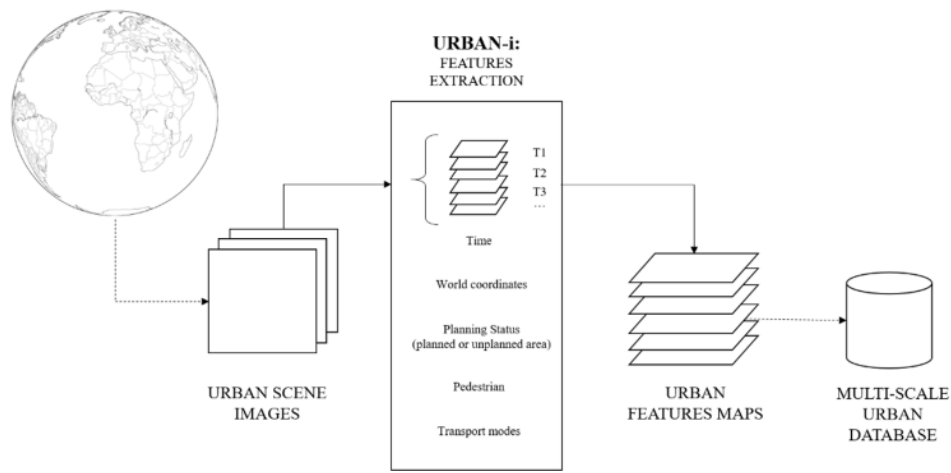

Fig. 2
URBAN-I CONCEPT

In this research, we introduce a novel multi-purpose urban modelling technique that offers a realistic environment for understanding cities. The architecture of URBAN-i relies on both deep learning and computer vision. It consists of two deep Convolutional Neural Networks (CNN), (LeCun, Bengio, & Hinton, 2015), as sub-models that are pre-trained differently on different data sets for accumulative purposes. URBAN-i aims to map the some of the agents of cities (pedestrian, transport modes, and settlement conditions) at a given time and space with the respect to the complexity of the urban settings. The goal of this model is not to geo-reference 'labelled images', but rather extract and geo-reference information from 'unlabelled urban scene images'. This will offer urban modellers a realistic platform for urban simulation for documenting city dynamics and tackling various urban issues. The current version of the model deals with information such as people, transport modes, and the planning status of the built-environment (planned region, or deteriorated or slums region) to bridge the gap in understanding cities in both developed and developing countries, including informal and slum regions.

Each urban scene image is a unique attribute for its location and time. Accordingly, the model geo-locate data captured from images according to their actual temporal scale (form an annual-scale to a second-scale). This will allow a precise understanding of the dynamics of the urban world for any urban scholar-defined research. We argue that URBAN-i can exemplify the possibility of moving from pseudo-dynamic modelling to realistic-dynamic modelling system with unified input data that are accessible by anyone in the everyday urban scene. For instance, by feeding the model enough images of a certain location, this will allow recalling the number of people at a given time and in a given space.

This research aims to exemplify the application of computer vision and deep learning in understanding city development beyond the concept of smart cities that are discussed and perceived widely in the mainstream. The novelty of the proposed model appears in applying computer vision and deep learning to understand the intangible and qualitative measures of the built environment such as the structure of informal settlements and slums. It offers an unprecedented opportunity for documenting the changes and dynamics of cities (See Fig. 1and 2).

*URBAN-i key features and highlights*

The URBAN-i model can be summarized as a computer vision tool for multipurpose urban modelling that can be used for data visualization or plugin any statistical, deep learning model to carry-on further analysis or prediction on acquired data. However, the key features of this version of the model that can be performed at a local, or on a global scale are:
1. Mapping slums, and deteriorated regions in cities,
2. Documenting the dynamics and changes in cities (with a precision of a second),
3. An actual environment for urban simulation and agent-based modelling,
4. Automated data extraction from cities for data visualization, or any researcher defined model for analysis,
5. Modelling transport modes and pedestrian flows.

The research is structured as follows: section 2 gives a brief

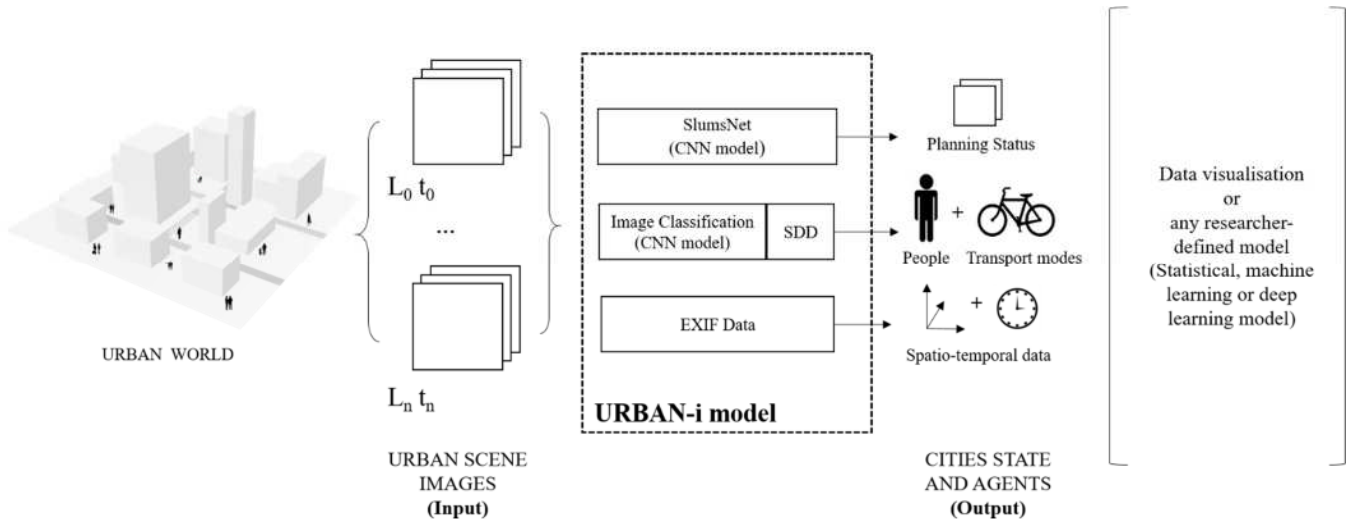
FIG. 3
URBAN-I ARCHITECTURE

a literature review of the related work to the URBAN-i model. Section 3 describes the model architecture and methodology, which it is divided into three sections. First, we will discuss the training of the convolutional neural network to sense and detect the overall planning status of an image. Second, we will discuss the implementation of this model to the state of art object-based detection methods. Last, we will use the previous two models for urban modelling and mapping and extract features to maps. Section 4 shows the model results and validation. Section 5 describes the model advancement and limitation. And lastly, section 6 gives a summary and conclusion of the research, highlighting the future work of this model.

## 2. RELATED WORK

The research related to this research is rather in the field of computer science than in urban studies, despite its application to urban planning and the science of cities.

### 2.1. Image classification

Deep learning models, most specifically Convolutional Neural Networks (CNN), have shown substantial progress in classifying images of a wide spectrum of classes (LeCun, Bengio, & Hinton, 2015). Various deep CNN models with different architecture and hyper-parameters have been computed to recognize visual objects in a large repository of images, such as the ImageNET dataset that contains 15 million images that belong to 22,000 different classes (Russakovsky et al., 2015, 2015). Starting with AlexNet (Krizhevsky, Sutskever, & Hinton, 2012), VGGNet (Simonyan & Zisserman, 2014), GoogLeNet (Szegedy, Liu, Jia, Sermanet, & Reed, 2015), ResNets (He, Zhang, Ren, & Sun, 2015) and most recently, DenseNet (Huang, Liu, Weinberger, & van der Maaten, 2017), the accuracy of these CNN models have reached a significant improvement to recognize and classify a wide range of images.

### 2.2 Segmentation and object-based detection

Building on the task of image classification, various techniques using CNN models has emerged not only to recognize different image classes but rather to localize and detect multi-objects in a single image. Moving from Region based CNN model (R-CNN) (Girshick, Donahue, Darrell, & Malik, 2014), Fast R-CNN (Ren, He, Girshick, & Sun, 2016), to MultiBox Detectors for fast image segmentation, or so-called; Single Shot MulitBox Detector (SSD) technique (Liu et al., 2016), again CNN models have shown a significant impact on recognizing and detecting multi-objects in images with a good accuracy and speed.

On the other hand, understanding the different components of an urban scene from perspective images relying on computer vision was seminal for various applications for smart cities and self-driving cars (Li et al., 2017). Scene parsing relying on semantic segmentation is a continuous success of CNN models for understanding and classifying the different components of an urban scene (Badrinarayanan, Kendall, & Cipolla, 2016; Chen, Papandreou, Kokkinos, Murphy, & Yuille, 2016; Chen, Papandreou, Schroff, & Adam, 2017; G. Lin, Milan, Shen, & Reid, 2017; Long, Shelhamer, & Darrell, 2015; Peng, Zhang, Yu, Luo, & Sun, 2017; Yu & Koltun, 2015; Zhao, Shi, Qi, Wang, & Jia, 2017). This pixel-level classification made it possible to recognize and understand the deep subtleties of the different components of an urban scene (i.e. road area, building, people, cars, vegetation). While such a complex approach is still exclusive towards the applications of autonomous vehicles, it can be used to understand and extract information for urban studies and urban modelling.

### 2.3 City perception and sensing the overall

Unlike image classification, segmentation, and object recognition, understanding the overall gist of a scene is seminal for understanding cities (Oliva & Torralba, 2006), where few works have been done in this area for better understanding urban areas. For instance, sensing that qualitative measures that are underlined in urban scenes, such as the safety of a certain region, or even deterioration and poverty of a certain neighbourhood in comparison to their surrounding ones. The complexity of tackling this subject appears in the training phases of any deep learning model, nevertheless, in finding the ground truth that can hold true for different regions and cities.

Most significant work related to this subject is the quantification of the perception and appearances of streetscapes

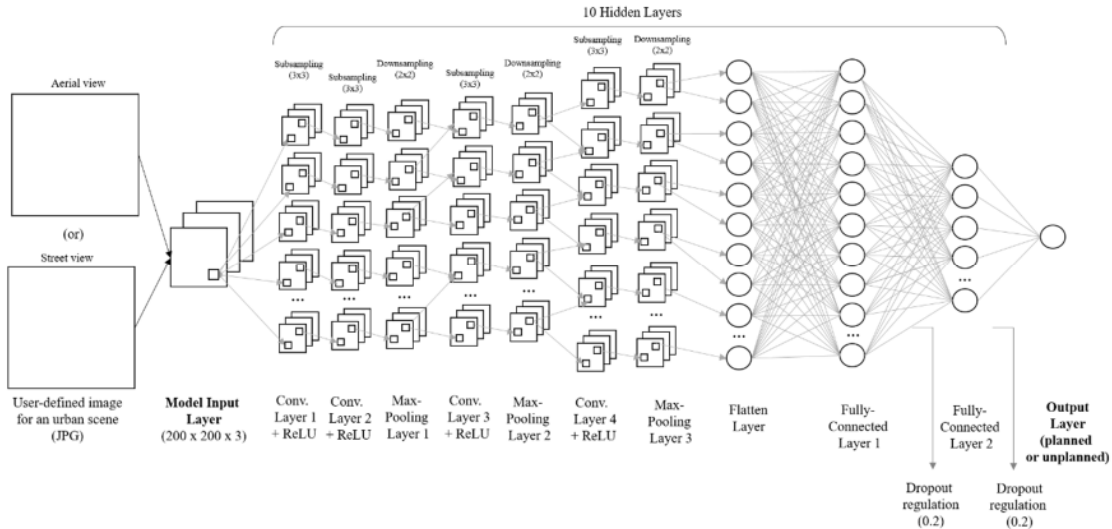

FIG. 4
THE ARCHITECTURE OF THE DEEP CNN MODEL

(Dubey, Naik, Parikh, Raskar, & Hidalgo, 2016; Naik, Raskar, & Hidalgo, 2016). This model has been conducted to detect humans' judgments concerning the quality of a region in a city that is collected via crowdsourcing. The proposed model relying on Support vector machine method has shown a significant impact on how machine learning models can learn and detect the perception of an urban area. Also, Salesses, Schechtner, & Hidalg (2013) have used urban scene images to map the inequality of urban perception.

## 3. METHODOLOGY

The architecture of URBAN-i model is divided into three sections: First, the SlumsNet model, second: the object-based detection model, and last, spatiotemporal data extraction. Such methods will allow the extraction of the planning status, agents- such as people and transport modes -from any urban scene image.

The extracted data will be geo-referenced based on their location and time. Fig.3 illustrates the overall methodology and architecture of the URBAN-I model. The complexity of this model is in the training and defining the algorithm, however for usage, it is a user-friendly model. For a given directory of images, the model will produce a .CSV file of data detected in each image. These data are latitude, longitude, year, month, day, hour, minute, second, planning status, person, car, bus, motorbike.

The URBAN-i model is fully-coded in Python programming, using mainly three deep learning libraries; TensorFlow, Keras, and Pytorch. Its training may take up to several days of computation on a computer with a normal CPU. For instance, training the current version of the SlumsNet model have taken 24 hours on a computer with i7 processor and 16 GB RAM. In the supplementary data section, we have attached the pre-trained model to be used for predictions on user-defined images.

### 3.1. Slums and deteriorated regions recognition

The SlumsNet is a new deep CNN model that is trained to sense and detect the differences between planned or unplanned urban scenes, including deteriorated and slum regions. This sub-model aims to understand the deep nuances of urban scenes according to their planning status; either formally planned, or informal regions. What makes this classification a complex process is that the task is not based on object detection but rather the deep understanding of the overall urban scene. Urban scenes consist of a wide spectrum of urban components that similarly belongs to planned or unplanned areas (such as people, buildings, materials, vegetation, water features, urban design elements ...etc.). Accordingly, the model has only to sense and understand the overall chaos behind the unplanned or the deteriorated areas in comparison to the formally planned ones, nevertheless, extract features that represents the chaos of informality, regardless to the similarities of these urban components or their counts and densities that may or may not exist in an urban scene.

*3.1.1 SlumsNet architecture*

We have built a deep Convolutional Neural Network (CNN) model (Guo et al., 2016; Hinton, Osindero, & Teh, 2006; LeCun, Bengio, & Hinton, 2015). The model transforms these images into an input layer of a size (200 x 200 x 3). The selection of the input resolution is based on trial and error in the sense that would enhance the model accuracy of prediction while optimising the required computation power to train the model.

Fig. 4 illustrates the overall architecture of the SlumsNet model. It is built based on 10 hidden layers of different types. The intuition for selecting the types and the orders of this network is based on trial and error for minimising the model loss, nevertheless, the knowledge transfer and understanding the intuition behind previous deep CNN models that are used for image classification tasks, such as VGGNet (Simonyan & Zisserman, 2014).

After the input layer, the model consists of 4 Convolutional layers, after the first two, each layer is followed by a Max-pooling layer. After the Flatten layer, two fully-connected layers are applied, followed by a single neuron output layer. The First three layers consist of 32 Feature maps of subsampling (3 x 3), While the third one consists of 128 feature maps of subsampling (3 x 3). The four convolutional layers rely on

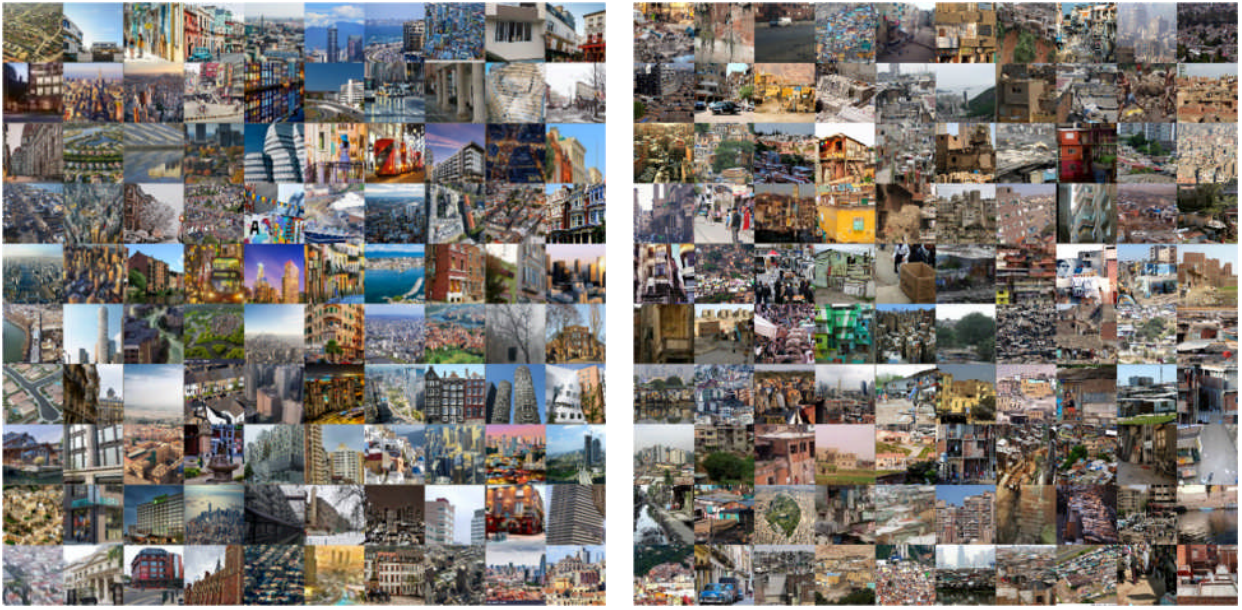

FIG. 5
A RANDOM SAMPLE OF URBAN SCENES OF THE PLANNED (RIGHT SIDE) AND THE UNPLANNED REGIONS (LEFT SIDE)

Rectified Linear Unit (ReLU) as an activation function to increase the nonlinearity of the model and enhance the performance of the neurons (Dahl, Sainath, & Hinton, 2013; Glorot, Bordes, & Bengio, 2011). It is defined as:
$$f(x) = \max(0, x) \quad (1)$$
where $x$ is the input neuron.

Moreover, the three Max-pooling layers are of a downsampling size (2 x 2). These layers are responsible for reducing dimensionality, in addition, it allows the model to adapt to the variation of the scale, rotation, or skewing of samples that represent a certain feature (Scherer, Müller, & Behnke, 2010). After these convolutional and Max-pooling layers, the flatten layers allow the model to convert the feature maps to neuron vectors that can be fed forward to the two fully-connected layers. The first fully-connected layer consists of 256 neurons, whereas the second one consists of 64 neurons. Both of these layers are activated based on a ReLU function. In order to avoid overfitting, we have applied feature dropout regulation after several hidden layers (Dahl et al., 2013; Srivastava, Hinton, Krizhevsky, Sutskever, & Salakhutdinov, 2014).

The output layer is based on a binary output of a single neuron, which classifies whether the image belongs to the planned or unplanned region. It is activated based on a sigmoid function that is defined as:
$$\delta(x) = \frac{1}{1+e^{-x}} \quad (2)$$
where $x$ is the input neuron.

The model is trained using back-propagation of error to update the weights of the neurons of a batch size of 32. It is compiled based on the optimization algorithm of stochastic gradient descent, relying on 'adam' optimizer (Kingma & Ba, 2014). The model is trained by 3 epochs; each consists of 9000 steps for training and 2000 steps for validation. The accuracy of the model is based on the cost function of Cross-Entropy error; in which it is defined as:
$$E = -\sum_i^n t_i \log(y_i) \quad (3)$$
where $t_i$ is the target vector, $y_i$ is the output vector, n represents the number of classes.

*3.1.2 Data*

There are no existing deep learning datasets that label and classify the built-up environment based on their planning status. Therefore, creating our own dataset was the only way to conduct this deep learning model. The data used for training and testing consists of 3000 images downloaded randomly from the internet that evenly belongs to the two groups; planned area, or unplanned area including informal, slum or deteriorated areas. These images belong to urban scenes that are taken from across the globe.

The ground truth for the model is defined based on three criteria. First, the obvious case of the deterioration of the region from where the urban scene image belongs. Second, the current literature of the locations in cities that belongs to slums or informal regions. Third, the metadata associated with the images from search engines (such as Google search engine) when data is gathered. Put all together, the collected images are assessed for ensuring label relevancy and a wide representation of image orientation and lighting conditions by visual inspection and only then the images are labelled to either formal or informal area, including slums and deteriorated regions. It is worth mentioning that these images are only used for training and validation, whereas the images presented in the results section for further validation are taken by the authors.

The CNN model is trained on 80% of the data set 2400 images) and tested for validation on the remaining 20% (600 images). Fig. 5 shows a random sample of the training and testing dataset for planned and unplanned areas respectively.
In order to allow a higher degree of freedom for analysing the status of a wide spectrum of urban scene images, we trained the model to understand both aerial perspective and street view images. Accordingly, this will allow the model to identify the status of the urban scenes, regardless of the angle and the elevation of the input image.

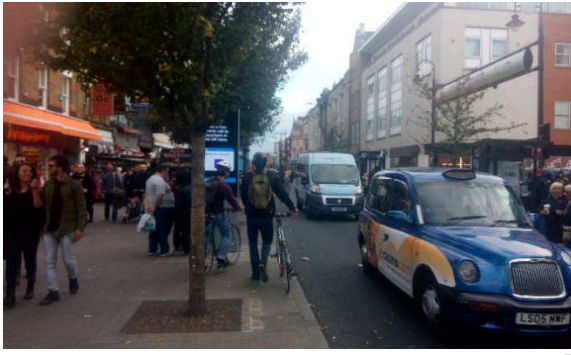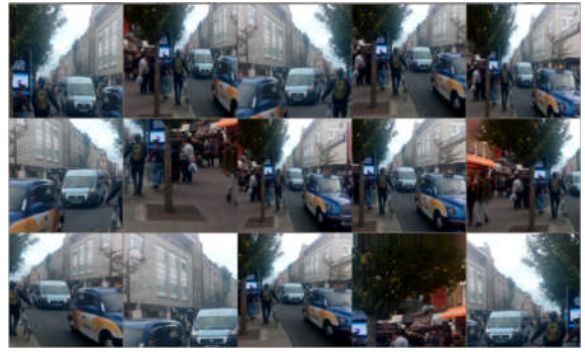

FIG. 6
THE URBAN SCENE FOR A PLANNED AREA IN LONDON AND EXAMPLES OF DATA AUGMENTATION
SOURCE: IMAGE CAPTURED BY THE AUTHOR

The images used for training are still limited. According, we have applied a data augmentation technique to enhance the training of the model. The algorithm allows the model to create random images based on four attributes; rescale, shear, zoom, and horizontal flips. While these approaches augment the training data, yet they do not change the class of the images, nonetheless, they offer realistic possibilities that can represent various urban scenes of the same location. Fig. 6 shows an example of an original image and a sample of the augmented data generated from the original image.

*3.2. Object-based detection sub-model*

*3.2.1 Model architecture*

To detect people and transport modes from urban scenes, we have used a Single Shot Mulibox Detector (SSD) method (Liu et al., 2016). Unlike other object detection approaches, SDD relies on a single feed-forward deep CNN model. It produces bounding boxes and a confidence score for each category of objects presented in the image. There are three reasons for selecting this approach for object detection. First, the model relies on a single deep CNN model to do the prediction and this makes it easier and faster to train. Second, this state-of-the-art method for object detection shows competitive results when it compared to the other object detection methods in many deep learning datasets, such as PASCAL VOC2007 (Everingham et al., 2015), COCO (Lin et al., 2014), and ILSVRC. Last, the model performs a fast real-time detection that makes it a good approach for URBAN-i to model the dynamics of the urban systems.

The model architecture relies on a base network for high-quality image classification, as discussed in section 2, that is truncated before the layers of classification, and an additional structure is added to the network. The first base of this network is built on the architecture of the VGG16 model that deals with classifying several image categories (Simonyan & Zisserman, 2014), including people and the different transport modes. The second part of the model relies on multi-scale feature maps and convolutional predictors for object-detection. These added convolutional features enable the model to detect an object at different scales and gives a confidence score for each bounding box for the occurrence of an object in the image.

The major difference of this approach in comparison to the training of other detectors is that the model only requires an input image with a bounding box as a ground truth. This facilitates the training process of the model while maintaining a high accuracy for object detection.

The objective loss function of the model is defined based on the weighted sum of the confidence loss (conf) and the localization loss (loc) (Liu et al., 2016). It is computed as:

$$L(x,l,g) = \frac{1}{N}\left(L_{conf}(x,c) + \alpha\, L_{loc}(x,l,g)\right) \quad (4)$$

where $N$ is the number of the matched default bounding boxes, if $N = 0$, the loss is set to 0, $\alpha$ is set to 1 by cross-validating the model. The confidence loss ($L_{conf}$) is defined based on a softmax loss for the confidence of the different classes (c).

$$L_{conf}(x,c) = -\sum_{i \in Pos}^{N} x_{ij}^p \log(\hat{c}_i^p) - \sum_{i \in Neg} \log(\hat{c}_i^0),$$

where $\hat{c}_i^p = \frac{\exp(c_i^p)}{\sum_p \exp(c_i^p)} \quad (5)$

The localization loss ($L_{loc}$) is a smooth loss between the parameters of the predicted box ($l$) and the ground truth bounding box ($g$) where the centre of the default bounding box (d) is ($cx, xy$) and its width ($w$) and height ($h$). It is computed as:

$$L_{loc}(x,l,g) = \sum_{i \in Pos}^{N} \sum_{m \in \{cx,xy,w,h\}} x_{ij}^k\, smooth\, L_1(l_i^m - \hat{g}_i^m) \quad (6)$$

We have implemented the SSD technique using the Pytorch library in Python programming (deGroot & Brown, 2017).

*3.2.2 Training Data for object detection*

The weights of the base network of the model, VGG16 model, has been trained on ILSVRC CLS-LOC dataset (Russakovsky et al., 2015). This gives us the opportunity to adopt it and extract these two main components for URBAN-i. After truncating the base network by converting the last fully connected layers to convolutional layers, and adapting its network with pre-discussed changes, the model is trained on PASCAL VOC 2007 dataset for image recognition (Everingham et al., 2015), with an initial learning rate of 0.1, and a momentum of 0.9. This dataset contains various visual object classes that are captured in realistic scenes, in which it is used primarily for supervised object detection. Its classes include a person, animal, vehicle type, and indoor objects.

For strengthening the model performance for different object size, a data augmentation technique has been used. For each image, the model computes several random samplings based on various techniques, such as sampling the patch in a way that the minimum Jaccard overlapping the object is either numerically defined or randomly sampled.

*3.3. Spatio-temporal data extraction*

This sub-model deals with the extraction of the coordinates and the time data of where and when the urban scene images are taken. This information is extracted from the Exchangeable image file format (EXIF) data that is accompanied by an image that features GPS data. This will allow the model to capture changes of the urban world according to a wide range of temporal scales, from the scale of year to even a second, to cope with the nature and the interdisciplinary of urban modelling tasks.

The algorithms define different functions to extract the coordinates, date and time, where the URBAN-i model can iterate through images to identify and extract these data and write it to a file besides the data captured from the SlumsNet and the object detection sub-models.

In order to not only extract the geographical coordinates data but also to extract time and date data, we have defined three functions that deal with each task separately. First, we have defined a function to extract the X and Y coordinates and convert them into the latitude and longitude. Second, we have defined an array to extract date (year, month, and day), Last, we have defined an array to extract time (hour, minute, second). The Python code is adapted and modified based on Sandler (2011).

4. RESULTS

*4.1 SlumsNet*

After training the SlumsNet to detect the planning status of the built-up environment, the validation accuracy of the testing data set is 85%. As we aim to use the model as a pragmatic tool for mapping slum regions, here we present a few examples of prediction of single images of the various urban scene taking from different cities across the globe.

Fig. 7 shows the prediction and the ground truth for some example of urban scene images, as a step forward to verifying the model further as a tool that can be used by planners and policy-makers across the globe. It shows a wide range of images that the model has predicted correctly and incorrectly for both classes. It is worth mentioning that the model can also classify images of a day-time or night-time shots of different weathers, including sunny, foggy, rainy, or snowy weathers. Nonetheless, the age of buildings is another crucial issue. In many cases, there are newly built buildings that belong to the unplanned area, whereas there are many historical buildings that belong to planned areas. While this variation complicates the training process and adds limitations to the model, however, it allows the proposed model to be widely used and furtherly developed to meet various mapping or sensing purposes.

*4.2 SlumsNet + Object-based detection*

The pre-trained SDD model for object detection of resolution (300x300) shows an overall validation accuracy of 77.2% for all classes on the VOC2007 Test database. This makes the model reliable to be used as a pragmatic tool. Fig. 8 shows the prediction of the URBAN-i model for the two trained deep CNN models. By adjoining the SlumsNet model with the object-based detection model, we enable urban modellers, planners and policy-makers to tackle the realistic dynamics of cities. By inputting urban scene images to the model, this model can allow urban modellers to carry-on the task of mimicking the actual setting and processes of complex urban systems without the need for oversimplifying the initial setting of the built environment. Accordingly, we argue that URBAN-i can exemplify the possibility of moving from pseudo-dynamic modelling, (Batty, 1984), to authentic-dynamic modelling system with minimal input data that are accessible in the everyday urban scene. This will allow mapping of the dynamics of the planning status of the region, including modelling pedestrian and transport modes.

*4.3 Mapping the world by the URBAN-i model*

Put all the algorithms of the URBAN-i model together, Fig. 9 shows the significant impact of the model of extracting information from urban scene images to a database that can be used for various urban research and data visualisation to understand the dynamics of a region, city or even an entire country. This database consists of information regarding the occurrence of transport modes, pedestrian, and the status of the planned area, in addition to the data of time and X-Y coordinates. Fig. 8 shows an example of urban scene images taken across the globe where the data is automatically geo-located and extracted to a database by the URBAN-i model. By the means of crowdsourcing and uploading images to a platform, the model directory can potentially include various images across the globe. This will enable urban modellers and researchers to collect and analyse their data sets based on the needs of their research.

5. DISCUSSION AND LIMITATIONS

In this article, we aimed to contribute to the advancements in the methods of urban modelling to better understand cities by using deep learning and computer vision. The URBAN-i model offers the opportunity of analysing urban scene images and mapping the planning status of an urban scene while mapping the occurrence of transport modes and pedestrians. The model aims to provide a potential opportunity to map an entire city by only walking around while taking photos using a phone device camera and feeding those images to the model for creating a spatiotemporal dataset that could be suitable for various research purposes, such as mapping, visualisation and prediction.

The model precision in detecting and classifying the urban scenes depends on several factors. First, the individual accuracy of each pre-trained CNN model is a key factor. Each one can be fine-tuned to achieve better accuracy and results with larger training datasets, higher computational power, and deeper network that are very limited to the author. However, the goal of this research is not to optimise accuracy from a computer science perspective, but rather show evidence that the complexity of global urban issues such as mapping slums and deteriorated regions in cities, traffic congestions and crowd-mapping can be tackled by deep learning and computer vision with less effort and minimum data that are available and accessible by everyone anywhere in the globe, without the without the means of expensive sensors. Second, the model precision also relies on the accuracy of the GPS data provider.

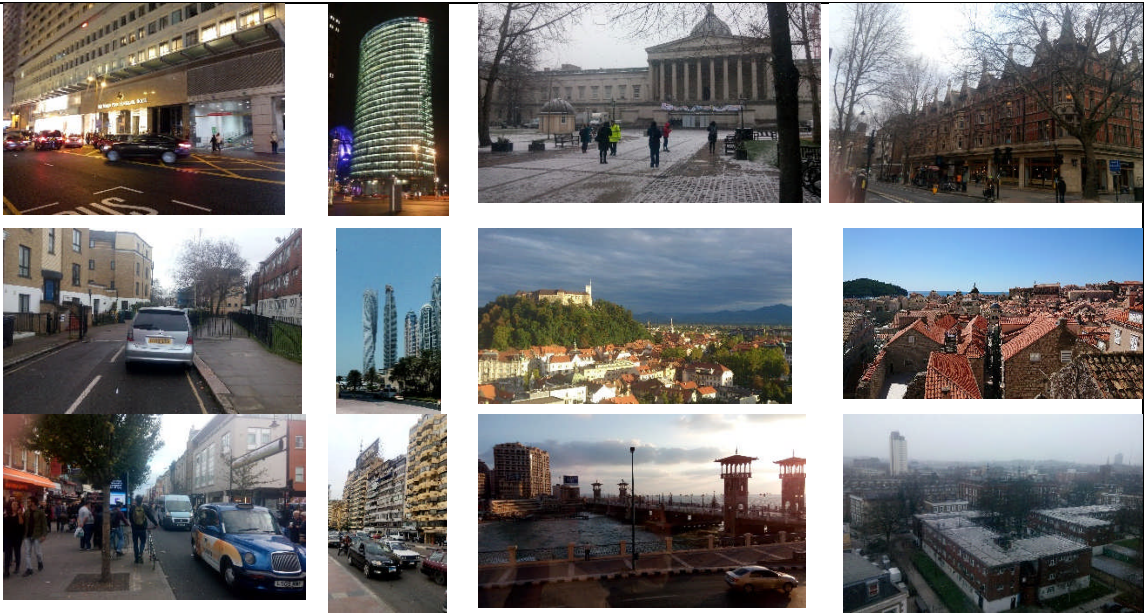
**Model correct prediction: Planned area**

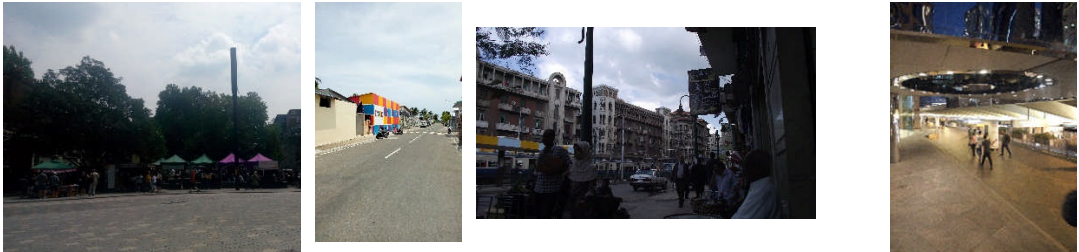
**Model incorrect prediction: Unplanned or deteriorated area**

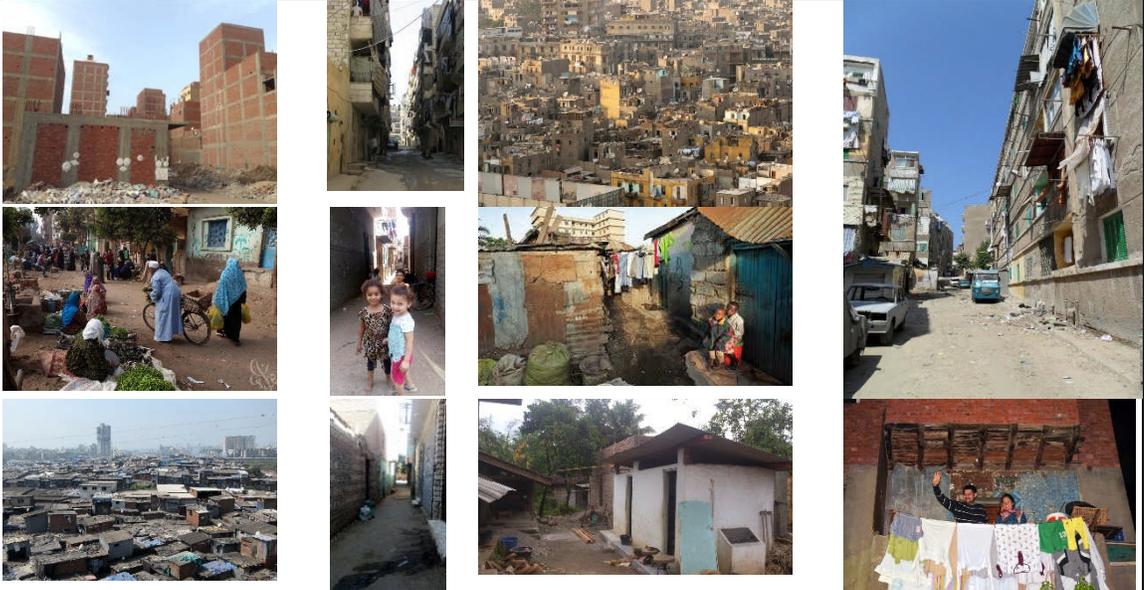
**Model prediction: Unplanned or deteriorated area**

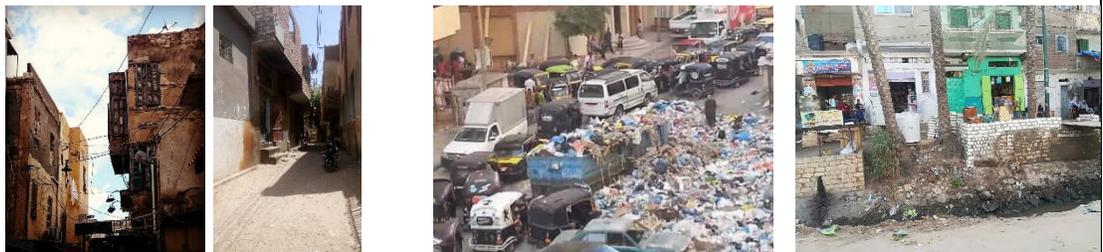
**Model incorrect prediction: Planned area**

Fig. 7
SlumsNet prediction in a sample of urban scene images

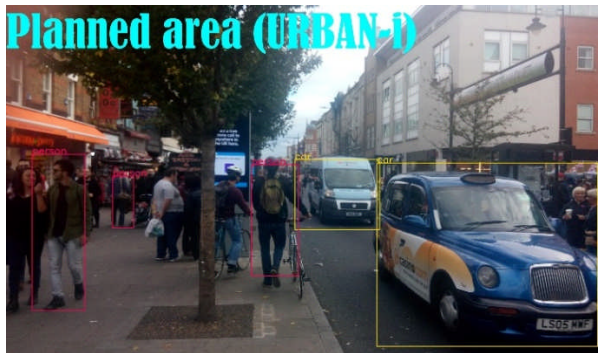
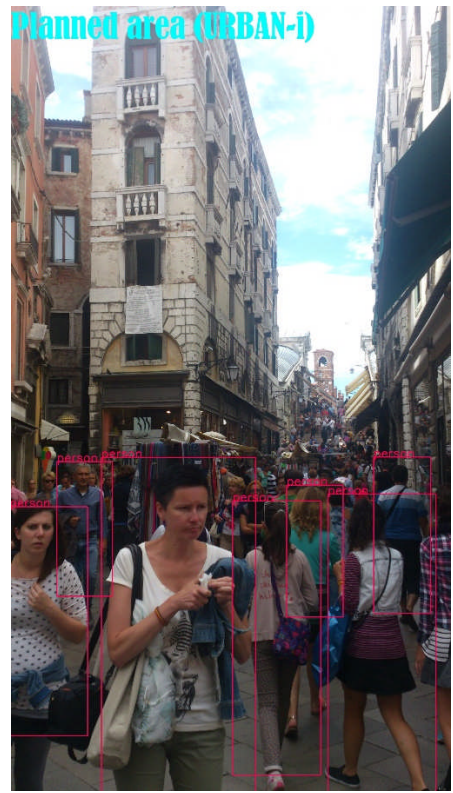
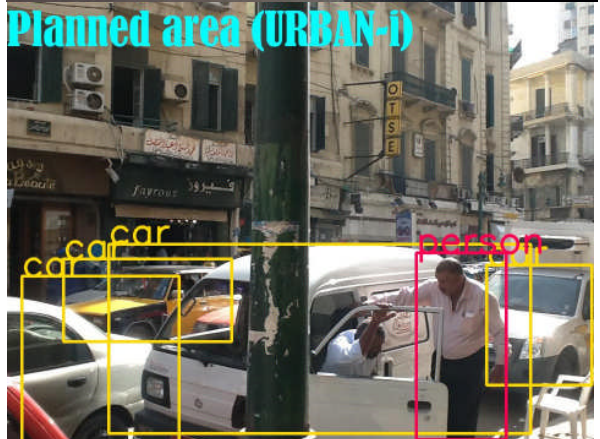
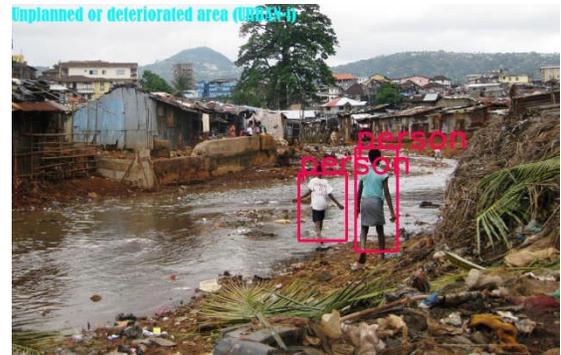
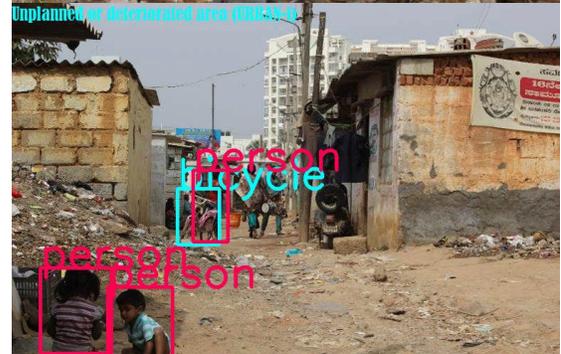
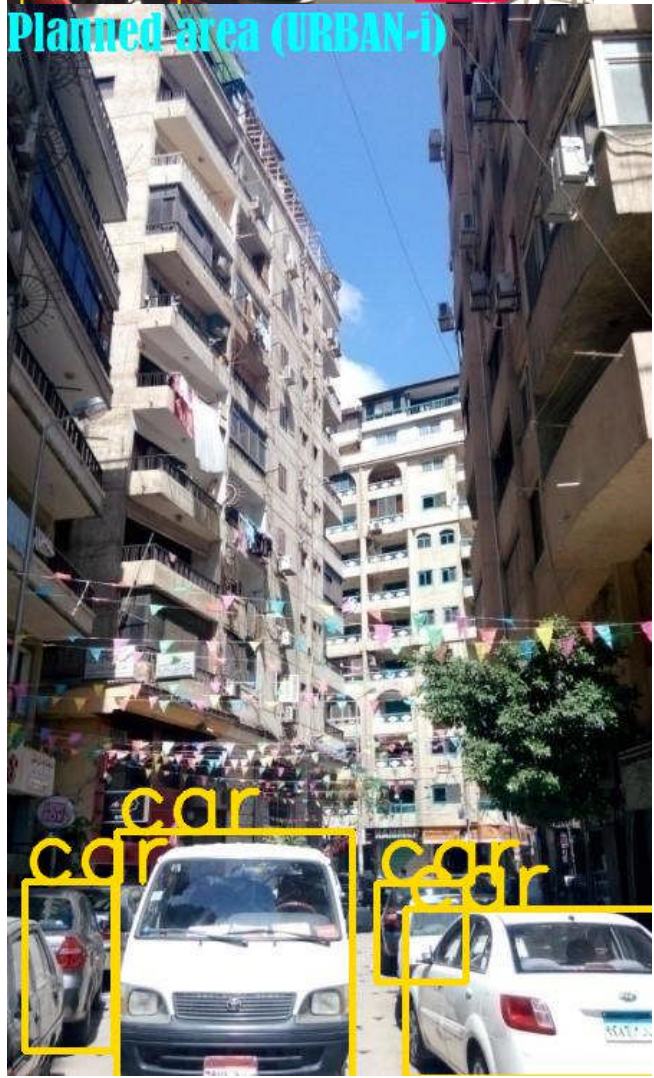
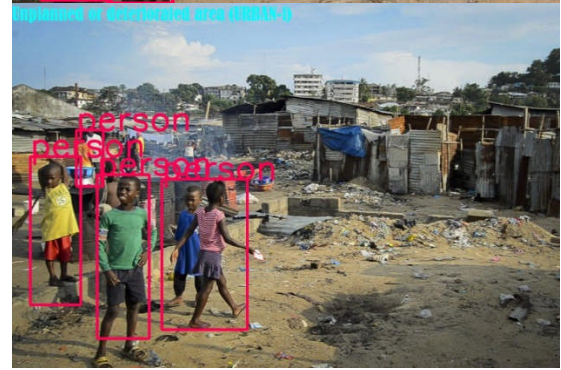

Fig. 8
SlumsNet and object-based detection in urban scene images

FIG. 9
URBAN-I DIRECTORY: FROM URBAN IMAGES TO THE AUTOMATED GLOBAL DATABASE

As for mapping a sequence of images, the accuracies of location data will be influenced by the GPS data receiver and transmitter when updating location data specifically when using the model for detection of rapid changes while changing locations. Last, the camera lens and shutter speed could also be another crucial factor when modelling fast-dynamics in cities. However, apart from these technological aspects that the model precision may be affected with it, the model shows significant potential for modelling complex events in everyday urban scenes within few seconds of detection, assuming that the images are instantly uploaded.

While the model shows novelty in analysing a wide range of images that belong to different planning status in cities, the model limitation appears in analysing images that are mixed with formal and informal housing in the same scene. A potential way to develop the model further is by using semantic segmentation and scene parsing. This pixel level segmentation would allow the model to provide a multiple categorization and localisation of the planning status for a single image. Accordingly, this will enhance the accuracy of the model when detecting complex scenes in the real world.

## 6. REMARKS AND FUTURE WORK

In this research, we presented a new computer vision model that can be utilized for various purposes of modelling the dynamics of the urban area at multi-scales; local and global. We have shown the possibility of mapping cities and tackling global urban such as informal area and slums recognition, traffic, and pedestrian modelling by using images taken from a mobile phone. This paper exemplifies the application of artificial intelligence and deep learning in understanding city development beyond the concept of smart cities that are discussed and perceived widely in the mainstream. The novelty of the proposed model appears in applying computer vision and deep learning to understand the intangible and qualitative measures of the built environment.

After training a deep CNN model, the SlumsNet, to understand the differences between images of a wide spectrum of planned and unplanned areas from across the globe, we obtained a validation accuracy of 85%. Such a high accuracy shows that even the chaos of informality in cities can be classified and detected relying on computer vision. Accordingly, we provide the model as a tool that can be used for mapping purposes or further developed to be used as a sensor for autonomous vehicles to understand the variation of the built-up environment when introducing such vehicles in less developed countries. Another potential application for the model is to be developed further to includes augmented reality information that can provide city dwellers or tourists with more information about the built environment.

The URBAN-i model is a multipurpose model that can be used for various tasks related to urban modelling. The current version of the model can be used for mapping the occurrence of transport modes, pedestrian and planning status of urban scenes in cities. Nevertheless, the model can be developed further to be used for modelling the dynamics of traffic congestion, crowd, even the deterioration, or the improvement of a region (planning status). Therefore, better decisions can be taken by policy-makers and planners to optimize resources and improve the living conditions in the urban world.

The model is fully-coded in Python programming using Anaconda to be used as a tool to capture and understand cities in different corners of the globe, including informal settlements and slum regions.

ACKNOWLEDGMENT

This research outcome is a part of a PhD study for the first author at University College London, supported by a partial fund of UCL Overseas Research Scholarship (ORS). We would like to thank Prof. Mike Batty for his feedback and discussion towards improving this article.

REFERENCES

[1] Baatz, G., Saurer, O., Köser, K., & Pollefeys, M. (2012). Large scale visual geo-localization of images in mountainous terrain. In *Computer Vision–ECCV 2012* (pp. 517–530). Springer.

[2] Badrinarayanan, V., Kendall, A., & Cipolla, R. (2016). SegNet: A Deep Convolutional Encoder-Decoder Architecture for Image Segmentation. arXiv:1511.00561v3 [cs.CV].

[3] Batty, M. (1984). *Pseudo-Dynamic Urban Models*. University of Wales Institue of Science and Technology. Retrieved from http://www.casa.ucl.ac.uk/PhD.pdf

[4] Batty, M. (2008). The Size, Scale, and Shape of Cities. *Science*, *319*(5864), 769–771. https://doi.org/10.1126/science.1151419

[5] Batty, M., & Longley, P. (1994). *Fractal Cities: A Geometry of Form and Function*. New York: Academic Press.

[6] Batty, M, & Xie, Y. (1996). Preliminary Evidence for a Theory of the Fractal City. *Environment and Planning A*, *28*(10), 1745–1762. https://doi.org/10.1068/a281745

[7] Batty, Michael. (1997). Cellular automata and urban form: a primer. *Journal of the American Planning Association*, *63*(2), 266–274.

[8] Batty, Michael. (2005). Agents, Cells, and Cities: New Representational Models for Simulating Multiscale Urban Dynamics. *Environment and Planning A*, *37*(8), 1373–1394. https://doi.org/10.1068/a3784

[9] Batty, Michael, Couclelis, H., & Eichen, M. (1997). *Urban systems as cellular automata*. SAGE Publications Sage UK: London, England.

[10] Batty, Michael, & Torrens, P. M. (2001). Modelling complexity : The limits to prediction. *Cybergeo*. https://doi.org/10.4000/cybergeo.1035

[11] Bettencourt, Luís. (2013). The origins of scaling in cities. *Science*, *340*(6139), 1438–1441.

[12] Bettencourt, Luis, & West, G. (2010). A unified theory of urban living. *Nature*, *467*(7318), 912–913.

[13] Chen, L.-C., Papandreou, G., Kokkinos, I., Murphy, K., & Yuille, A. (2016). SEMANTIC IMAGE SEGMENTATION WITH DEEP CONVOLUTIONAL NETS AND FULLY CONNECTED CRFS.

[14] Chen, L.-C., Papandreou, G., Schroff, F., & Adam, H. (2017). Rethinking atrous convolution for semantic image segmentation. *ArXiv Preprint ArXiv:1706.05587*.

[15] Crandall, D. J., Backstrom, L., Huttenlocher, D., & Kleinberg, J. (2009). Mapping the world's photos. In *Proceedings of the 18th international conference on World wide web* (pp. 761–770). ACM.

[16] Dahl, G. E., Sainath, T. N., & Hinton, G. E. (2013). Improving deep neural networks for LVCSR using rectified linear units and dropout. In *Acoustics, Speech and Signal Processing (ICASSP), 2013 IEEE International Conference on* (pp. 8609–8613). IEEE.

[17] de Almeida, C. M., Batty, M., Monteiro, A. M. V., Câmara, G., Soares-Filho, B. S., Cerqueira, G. C., & Pennachin, C. L. (2003). Stochastic cellular automata modeling of urban land use dynamics: empirical development and estimation. *Computers, Environment and Urban Systems*, *27*(5), 481–509.

[18] deGroot, M., & Brown, E. (2017). A PyTorch Implementation of Single Shot MultiBox Detector. Retrieved from https://github.com/amdegroot/ssd.pytorch

[19] Dubey, A., Naik, N., Parikh, D., Raskar, R., & Hidalgo, C. A. (2016). Deep learning the city: Quantifying urban perception at a global scale. In *European Conference on Computer Vision* (pp. 196–212). Springer.

[20] Everingham, M., Eslami, S. M. A., Van Gool, L., Williams, C. K. I., Winn, J., & Zisserman, A. (2015). The Pascal Visual Object Classes Challenge: A Retrospective. *International Journal of Computer Vision*, *111*(1), 98–136. https://doi.org/10.1007/s11263-014-0733-5

[21] Frankhauser, P. (n.d.). The fractal approach. A new tool for the spatial analysis of urban agglomerations, 37.

[22] Gallagher, A., Joshi, D., Yu, J., & Luo, J. (2009). Geo-location inference from image content and user tags. In *Computer Vision and Pattern Recognition Workshops, 2009. CVPR Workshops 2009. IEEE Computer Society Conference on* (pp. 55–62). IEEE.

[23] García, C., Meana-Llorián, D., G-Bustelo, B. C., Lovelle, J. M., & Garcia-Fernandez, N. (2017). Midgar: Detection of people through computer vision in the Internet of Things scenarios to improve the security in Smart Cities, Smart Towns, and Smart Homes. *Future Generation Computer Systems*, *76*, 301–313. https://doi.org/10.1016/j.future.2016.12.033

[24] Girshick, R., Donahue, J., Darrell, T., & Malik, J. (2014). Rich feature hierarchies for accurate object detection and semantic segmentation. arXiv:1311.2524 [cs.CV]. Retrieved from https://arxiv.org/pdf/1311.2524.pdf

[25] Glorot, X., Bordes, A., & Bengio, Y. (2011). Deep sparse rectifier neural networks. In *Proceedings of the Fourteenth International Conference on Artificial Intelligence and Statistics* (pp. 315–323).

[26] Guo, Y., Liu, Y., Oerlemans, A., Lao, S., Wu, S., & Lew, M. S. (2016). Deep learning for visual understanding: A review. *Neurocomputing*, *187*, 27–48. https://doi.org/10.1016/j.neucom.2015.09.116

[27] Hays, J., & Efros, A. A. (2015). Large-Scale Image Geolocalization. In J. Choi & G. Friedland (Eds.), *Multimodal Location Estimation of Videos and Images* (pp. 41–62). Cham: Springer International Publishing. https://doi.org/10.1007/978-3-319-09861-6_3

[28] He, K., Zhang, X., Ren, S., & Sun, J. (2015). Deep Residual Learning for Image Recognition. *ArXiv:1512.03385v1*. Retrieved from https://arxiv.org/pdf/1512.03385.pdf

[29] Heppenstall, A. J., Crooks, A. T., See, L. M., & Batty, M. (Eds.). (2012). *Agent-Based Models of Geographical Systems*. Dordrecht: Springer Netherlands. https://doi.org/10.1007/978-90-481-8927-4

[30] Hinton, G. E., Osindero, S., & Teh, Y.-W. (2006). A fast learning algorithm for deep belief nets. *Neural Computation*, *18*(7), 1527–1554.

[31] Huang, G., Liu, Z., Weinberger, K. Q., & van der Maaten, L. (2017). Densely connected convolutional networks. In *Proceedings of the IEEE conference on computer vision and pattern recognition* (Vol. 1, p. 3).

[32] Isalgue, A., Coch, H., & Serra, R. (2007). Scaling laws and the modern city. *Physica A: Statistical Mechanics and Its Applications*, *382*(2), 643–649. https://doi.org/10.1016/j.physa.2007.04.019

[33] Kingma, D. P., & Ba, J. (2014). Adam: A method for stochastic optimization. *ArXiv Preprint ArXiv:1412.6980*.

[34] Krizhevsky, A., Sutskever, I., & Hinton, G. E. (2012). Imagenet classification with deep convolutional neural networks. In *Advances in neural information processing systems* (pp. 1097–1105).

[35] LeCun, Y., Bengio, Y., & Hinton, G. (2015). Deep learning. *Nature*, *521*(7553), 436–444. https://doi.org/10.1038/nature14539

[36] Li, X., Wang, Z. J. W., Yang, C. L. J., Chen, X. S. Z. L. Q., Yan, S., & Feng, J. (2017). FoveaNet: Perspective-aware Urban Scene Parsing. *ArXiv Preprint ArXiv:1708.02421*.

[37] Lin, G., Milan, A., Shen, C., & Reid, I. (2017). Refinenet: Multi-path refinement networks for high-resolution semantic segmentation. In *IEEE Conference on Computer Vision and Pattern Recognition (CVPR)*.

[38] Lin, T.-Y., Maire, M., Belongie, S., Bourdev, L., Girshick, R., Hays, J., … Dollár, P. (2014). Microsoft COCO: Common Objects in Context. *ArXiv:1405.0312 [Cs]*. Retrieved from http://arxiv.org/abs/1405.0312

[39] Liu, W., Anguelov, D., Erhan, D., Szegedy, C., Reed, S., Fu, C.-Y., & Berg, A. C. (2016). Ssd: Single shot multibox detector. In *European conference on computer vision* (pp. 21–37). Springer.

[40] Long, J., Shelhamer, E., & Darrell, T. (2015). Fully Convolutional Networks for Semantic Segmentation. arXiv:1411.4038v2 [cs.CV].

[41] Lynch, K. (1960). *The image of the city*. The technology press and Harvard University press.

[42] Murcio, R., Masucci, A. P., Arcaute, E., & Batty, M. (2015). Multifractal to monofractal evolution of the London street network. *Physical Review E*, *92*(6). https://doi.org/10.1103/PhysRevE.92.062130

[43] Naik, N., Raskar, R., & Hidalgo, C. A. (2016). Cities Are Physical Too: Using Computer Vision to Measure the Quality and Impact of Urban

Appearance. *American Economic Review*, *106*(5), 128–132. https://doi.org/10.1257/aer.p20161030

[44] Oliva, A., & Torralba, A. (2006). Chapter 2 Building the gist of a scene: the role of global image features in recognition. In *Progress in Brain Research* (Vol. 155, pp. 23–36). Elsevier. https://doi.org/10.1016/S0079-6123(06)55002-2

[45] Peng, C., Zhang, X., Yu, G., Luo, G., & Sun, J. (2017). Large Kernel Matters–Improve Semantic Segmentation by Global Convolutional Network. *ArXiv Preprint ArXiv:1703.02719*.

[46] Quack, T., Leibe, B., & Van Gool, L. (2008). World-scale mining of objects and events from community photo collections. In *Proceedings of the 2008 international conference on Content-based image and video retrieval* (pp. 47–56). ACM.

[47] Ren, S., He, K., Girshick, R., & Sun, J. (2016). Faster R-CNN: Towards Real-Time Object Detection with Region Proposal Networks. arXiv:1506.01497v3.

[48] Russakovsky, O., Deng, J., Su, H., Krause, J., Satheesh, S., Ma, S., … Fei-Fei, L. (2015). ImageNet Large Scale Visual Recognition Challenge. *International Journal of Computer Vision*, *115*(3), 211–252. https://doi.org/10.1007/s11263-015-0816-y

[49] Salesses, P., Schechtner, K., & Hidalgo, C. A. (2013). The Collaborative Image of The City: Mapping the Inequality of Urban Perception. *PLoS ONE*, *8*(7), e68400. https://doi.org/10.1371/journal.pone.0068400

[50] Sandler, E. (2011). *get_lat_lon_exif_pil.py*. Retrieved from https://gist.github.com/erans/983821

[51] Scherer, D., Müller, A., & Behnke, S. (2010). Evaluation of pooling operations in convolutional architectures for object recognition. In *International conference on artificial neural networks* (pp. 92–101). Springer.

[52] Simonyan, K., & Zisserman, A. (2014). Very deep convolutional networks for large-scale image recognition. *ArXiv Preprint ArXiv:1409.1556*.

[53] Srivastava, N., Hinton, G., Krizhevsky, A., Sutskever, I., & Salakhutdinov, R. (2014). Dropout: A simple way to prevent neural networks from overfitting. *The Journal of Machine Learning Research*, *15*(1), 1929–1958.

[54] Szegedy, C., Liu, W., Jia, Y., Sermanet, P., & Reed, S. (2015). Going Deeper with Convolutions. Computer Vision Foundation. Retrieved from https://www.cs.unc.edu/~wliu/papers/GoogLeNet.pdf

[55] Tarigan, J., Nadia, Diedan, R., & Suryana, Y. (2017). Plate Recognition Using Backpropagation Neural Network and Genetic Algorithm. *Procedia Computer Science*, *116*, 365–372. https://doi.org/10.1016/j.procs.2017.10.068

[56] Yu, F., & Koltun, V. (2015). Multi-scale context aggregation by dilated convolutions. *ArXiv Preprint ArXiv:1511.07122*.

[57] Zhao, H., Shi, J., Qi, X., Wang, X., & Jia, J. (2017). Pyramid scene parsing network. In *IEEE Conf. on Computer Vision and Pattern Recognition (CVPR)* (pp. 2881–2890).

[58] Zhou, B., Liu, L., Oliva, A., & Torralba, A. (2014). Recognizing city identity via attribute analysis of geo-tagged images. In *European conference on computer vision* (pp. 519–534). Springer.